\documentclass{ecai2014}
\usepackage{times}
\usepackage{graphicx}
\usepackage{latexsym}
\usepackage{url}
\usepackage{tikz}
\usepackage{pdflscape}
\usetikzlibrary{shapes}

\begin{document}

\title{An Easy to Use Repository for Comparing and Improving Machine Learning Algorithm Usage}

\author{Michael R. Smith\institute{Brigham Young University, USA, email: msmith@axon.cs.byu.edu}
\and Andrew White\institute{Brigham Young University, USA, email: andrewkvavlewhite@gmail.com}
\and Christophe Giraud-Carrier\institute{Brigham Young University, USA, email: cgc@cs.byu.edu}
\and Tony Martinez\institute{Brigham Young University, USA, email: martinez@cs.byu.edu}}

\maketitle
\bibliographystyle{ecai2014}

\begin{abstract}
The results from most machine learning experiments are used for a specific purpose and then discarded.
This results in a significant loss of information and requires rerunning experiments to compare learning algorithms.
Often, this also requires a researcher or practitioner to implement another algorithm for comparison, that may not always be correctly implemented.
By storing the results from previous experiments, machine learning algorithms can be compared easily and the knowledge gained from them can be used to improve the performance of future machine learning experiments.
The purpose of this work is to provide easy access to previous experimental results for learning and comparison.
These stored results are comprehensive -- storing the prediction for each test instance as well as the learning algorithm, hyperparameters, and training set that were used in the experiment.
Previous experimental results are particularly important for meta-learning, which, in a broad sense, is the process of learning from previous machine learning results such that the learning process is improved.
While other experiment databases do exist, one of our focuses is on easy access to the data, eliminating any learning curve required to acquire the desired information.
We provide meta-learning data sets that are ready to be downloaded for meta-learning experiments.
Easy access to previous experimental results aids other researchers looking to do meta-learning and helps in comparing meta-learning algorithms.
In addition, queries to the underlying database can be made if specific information is desired.
We also differ from previous experiment databases in that our databases is designed at the instance level, where an instance is an example in a data set.
We store the predictions of a learning algorithm trained on a specific training set for each instance in the test set.
Data set level information can then be obtained by aggregating the results from the instances.
The instance level information can be used for many tasks such as determining the diversity of a classifier or algorithmically determining the optimal subset of training instances for a learning algorithm.
\end{abstract}

\section{INTRODUCTION}

The quality of an induced model is dependent on, among other aspects, the learning algorithm that is chosen, the hyper-parameter settings for the chosen learning algorithm, and the quality of the training set.
Choosing a learning algorithm for a given task, setting its hyper-parameters, and selecting which instances to train on, however, is non-trivial.
Meta-learning deals with the problem of how to select a learning algorithm and set its hyper-parameters based on previous experience (results from previous machine learning experiments).
Although some research from the machine learning community has focused on meta-learning (e.g., see~\cite{Pfahringer2000,Brazdil2003,Ali2006a,Ali2006b,Bergstra2012,Gomes2012}), much of the focus of machine learning research has been on developing more learning algorithms and/or applying machine learning in specific domains.

Part of the difficulty of meta-learning is due to the lack of accessible results.
As meta-learning requires running several learning algorithms and hyperparameter settings over a large set of data sets, gathering results requires large amounts of computational resources.
In addition to the computational requirements, results from the learning algorithms may differ due to slight differences in their implementations.
Thus, comparing results between meta-learning studies becomes difficult.

To aid in further research in meta-learning, we have developed the \textit{machine learning results repository} (MLRR) that provides data sets ready for download for meta-learning problems, akin to the UCI data repository for machine learning problems.
We refer to the data sets for meta-learning as \textit{meta-data sets} to distinguish them from the data sets that are used in the machine learning experiments.
The meta-data sets provide a snapshot of an underlying database that stores the results of machine learning experiments.
Users can update the database with new results from machine learning experiments and then update the meta-data sets for meta-learning.
A revision history is kept so that comparisons between meta-learning algorithms can be easily facilitated.
As a starting point, meta-data sets are provided by MLRR for typical meta-learning tasks such as given a set of meta-features, predict with learning algorithm and/or hyperparameter settings to use.

The MLRR stores instance level meta features and the predictions made on each instance by the learning algorithms.
Providing information at the instance level allows studies on the instances to be done.
Studying the effects of machine learning on a single instance and/or the effects of a single instance on the performance of an algorithm has generally been overlooked.
Instance-level information is important in several areas of machine learning.
In ensembles, computing the classifier diversity of the ensembled classifiers using the predictions for each instance is important in determining the effectiveness of the ensembling technique \cite{Kuncheva2003,Brown2005_JMLR,Aksela2006}.
Recently, curriculum learning incrementally augments the training set such that ``easier'' instances are presented to the learning algorithm first \cite{Bengio2009}.
Thus, creating a need to understand and identify the easier instances.
Smith et al. used instance-level predictions to identify and characterize instances that are likely to be misclassified \cite{Smith2012_IH} and used this information to create a curriculum \cite{Smith_CL}.
Other work also uses the instance-level predictions for meta-learning.
The classifier output difference (COD) \cite{Peterson2005} is a metric that measures the diversity between learning algorithms.
COD measures the distance between two learning algorithms as the probability that the learning algorithms make different predictions.
Unsupervised meta-learning (UML) clusters learning algorithms based on their COD scores (rather than accuracy) to examine the behavior of the learning algorithms \cite{Lee2011}.
Meta-learning can then be done over the clusters rather than a larger set of learning learnings to recommend a cluster of learning algorithms that all behave similarly \cite{Lee2013}.
Additionally, several techniques treat instances individually during the training process such as filtering instances from the training set based on their instance-level meta-features \cite{Smith2011} or weighting the instances \cite{Rebbapragada2007}. 


Other works have created a repository for machine learning experiments from which learning can be conducted \cite{Reif2012, Vanschoren2012} which either lacked simplicity and/or extensibility.
In addition to providing instance-level information, we hope to bridge that gap with the MLRR.
The most prominent and well-developed data repositories are experiment databases \cite{Vanschoren2012}, which provide a framework for reporting experimental results and the workflow of the experiment.
The purpose of the experiment databases is to comprehensively store the workflow process of the experiment for reproducibility.
One of the results of storing the experiments is that the results can be used for meta-learning.
Unfortunately, there is a learning curve to access the data due to the inherent complexity involved in storing all of the details about exact reproducibility.
Because of this complexity and formality, it is difficult to directly access the information that would be the most beneficial for meta-learning and may deter some potential users.
Additionally, currently it does not support any instance level features to be stored.

We acknowledge that maintaining a database of previous experiments is not a simple problem.
We add our voice to support the importance of maintaining a repository of machine learning results and offer an efficient possible solution for storing results from previous experiments.
Our primary goal is to maintain simplicity and provide easily accessible data for meta-learning to 1) help promote more research in meta-learning, 2) to provide a standard set of data sets for meta-learning algorithm comparison, and 3) to stimulate research at the instance-level.

We next describe our approach for providing a repository for machine learning meta-data the emphasizes ease of access to the meta-data.
MLRR currently has the results from 72 data sets, 9 learning algorithms and 10 hyperparameter settings for each learning algorithm.
The database description is then provided in Section \ref{section:DBdescription}.
How to add new experimental results to the database is detailed in Section \ref{section:extendingDB}.
We then give a more detailed description of the data set level and instance level meta-features that are used in the MLRR.
Conclusions and directions for future work are provided in Section \ref{section:conlcusions}.

\section{META-DATA SET DESCRIPTIONS}
\label{section:topDown}
The purpose of the \textit{machine learning results repository} (MLRR) is to provide easy access to the results from previous machine learning experiments for meta-learning at the data set and instance levels.
Thus, allowing other researchers interested in meta-learning and in better understanding machine learning algorithms direct access to prior results without having to re-run all of the algorithms or learn how to navigate a more complex experiment database. 
The quality of an induced model for a task is dependent on at least three items:
1) the learning algorithm chosen to induce the model,
2) the hyperparameter settings for the chosen learning algorithm, and
3) the instances used for training.
When we refer to an experiment, we mean the results from training a learning algorithm $l$ with hyperparamter settings $\lambda$ trained on the training set $t$.
We first describe how we manage experiment information.
We then describe the provided meta-data sets.

\subsection{Experiment information}
The information about each experiment is provided in three tables.
Which learning algorithm and hyperparameters were used is provided in a file structured as shown in Table \ref{table:LA_mapping}.
It provides the toolkit that was ran, the learning algorithm, and the hyperparameters that were used.
This allows for multiple learning algorithms, hyperparameters, and toolkits to be compared.
In the examples in Table \ref{table:LA_mapping}, the class names from the Weka machine learning toolkit \cite{weka2009} and the Waffles machine learning toolkit \cite{Gashler2011} are shown.
LA\_seed corresponds to the learning algorithm that was used (LA) and to a seed that represents which hyperparameter setting was used (seed).
The LA\_seed will be used in other tables to map back to this table. 
A seed of -1 represents the default hyperparameter settings as many studies examine the default behavior as given in a toolkit and the default parameters are commonly used in practice.

\begin{table}
\centering
{\caption{The structure of the meta-data set that describes the hyperparameter settings for the learning algorithms stored in the database.}\label{table:LA_mapping}}
\setlength{\tabcolsep}{2pt}
\begin{tabular}{l|l|l}
LA\_seed & Toolkit & Hyperparamters\\
\hline
\\[-6pt]
BP\_1 & weka & weka.classifiers.functions.MultilayerPerceptron\textbackslash\\&& -- -L 0.261703 -M 0.161703 -H 12 -D\\
BP\_2 & weka & weka.classifiers.functions.MultilayerPerceptron\textbackslash\\&& -- -L 0.25807 -M 0.15807 -H 4\\
BP\_3 & waffles & neuralnet -addlayer 8 -learningrate 0.1 -momentum 0\textbackslash\\&& -windowsepochs 50 \\
\vdots & \vdots & \vdots \\
C4.5\_1 & weka & weka.classifiers.trees.J48 -- -C 0.443973 -M 1 \\
\vdots & \vdots & \vdots \\
\end{tabular}
\end{table}

As the parameter values differ between toolkits, there is a mapping provided to distinguish hyperparameter settings.
For example, Weka uses the ``-L'' parameter to set the learning rate in backpropagation while the Waffles toolkit uses ``-learningrate''.
Also, some toolkits have hyperparameters that other implementations of the same learning algorithm do not include.
In such cases, an unknown value will be provided in the meta-data set.
This mapping is shown in Table \ref{table:hyperparamMapping} for the backpropagation learning algorithm.
The first row contains the values used by MLRR.
The following rows contain the command-line parameter supplied to a specific toolkit to set that hyperparameter.

\begin{table}
\centering
{\caption{The structure of the table for mapping learning algorithm hyperparameters between different toolkits for the backpropagation learning algorithms.}\label{table:hyperparamMapping}}
\setlength{\tabcolsep}{4pt}
\begin{tabular}{l|ccccc}
 & \multicolumn{5}{|c}{Command line parameters}\\
toolkit & LR & Mo & HN & DC & WE \\
\hline
\\[-6pt]
weka & -L & -M & -H & -D & ? \\
waffles & -learningrate& -momentum & -addlayer & ? & -windowsepochs \\
\vdots & \vdots & \vdots & \vdots & \vdots & \vdots\\
\end{tabular}
\end{table}

A mapping of which instances are used for training is also provided in a separate file.
The structure of this table is shown in Table \ref{table:foldMapping}.
Each row represents an experiment as toolkit\_seed\_numFolds\_fold.
The toolkit represents which toolkit was used, the seed represents the random seed that was provided to the toolkit, numFolds represents how many folds were ran, and fold represents in which fold an instance was included for testing.
The values in the following columns represent if an instance was used for training or testing.
There is one column for each instance in the data set.
They are stored as real values.
This allows for the cases where a training instance has an associated weight.
In the file, an unknown value of ``?'' represents a testing instance, otherwise a real-value represents a training instance.
A value of 0 represents a filtered instance, a 1 represents an unweighted training instance and any value between 0 and 1 represents the weight for that training instance.
In the cases where there are specific training and testing sets, then the row will be labeled as toolkit\_0\_0\_1 and information for the training set can be entered as before.
A random test/training split of the data is represented as toolkit\_seed\_percentSplit\_1 where ``percentSplit'' represents the percentage of the data set that was used for testing as generated by the tool kit.

\begin{table}
\centering
{\caption{The structure of the meta-data set that indicates which instances were used for training given a random seed.}\label{table:foldMapping}}
\begin{tabular}{c|cccc}
toolkit\_seed\_\# folds\_fold  & 1 & 2 & 3 & \dots\\
\hline
\\[-6pt]
weka\_1\_10\_1 & 1 & 1 & 1 & \dots\\
weka\_1\_10\_2 & 1 & 0 & 1 & \dots\\ 
\vdots & \vdots& \vdots& \vdots& \\
\\[-6pt]
weka\_1\_10\_10 & 0.74 & 1 & ? & \dots\\
weka\_2\_1\_10 & ? & 1 & 1 & \dots\\
\vdots & \vdots& \vdots& \vdots& \\
\end{tabular}
\end{table}

\subsection{Meta-data sets}

The MLRR is unique in the sense that it stores and presents instance level information, namely, instance level characteristics and the predictions from previous experiments.
From the instance level predictions, the data set metrics can be computed (e.g. accuracy or precision).
To our knowledge, the MLRR is the first attempt to store instance level information about the training instances.

As one of the purposes of the MLRR is ease of access, the MLRR stores several data sets in attribute-relation file format (ARFF) which is used in many machine learning toolkits.
In essence, ARFF is a comma or space separated file with attribute information and possible comments.
The precomputed meta-data sets include instance level meta-data sets and data set level meta-data sets.

At the instance level, MLRR provides for each data set a meta-data set that stores the instance level meta-features and the prediction from each experiment. 
This allows for analyses to be done exploring the effects of hyperparameters and learning algorithms at the instance-level, which is currently mostly overlooked.
For each data set, a meta-data set is provided that gives the values for the instance level meta-features, the actual class value (stored as a numeric value), and the predicted class value for each experiment.
The training set and learning algorithm/hyperparameter information is stored in the column heading as ``LA\_seed/hyperparameter'' where LA is a learning algorithm and hyperparameter is the hyperparameter setting for the learning algorithm.
Together, they map to the entries in Table \ref{table:LA_mapping}.
The seed represents the seed that was used to partition the data (see Table \ref{table:foldMapping}).
The structure of the instance level meta-data set is shown in Table \ref{table:instLevel}.
In the given example, instance 77 is shown.
The ``inst meta'' section provides the instance level meta-features for that instance.
The actual class label is label 2.
The predictions from the experiments on this data set are provided in the following columns (i.e. experiment BP\_1/1 predicted class 3, BP\_N/1 predicted class 2, etc.).

\begin{table}
\centering
{\caption{The structure of the meta-data set at the instance level.}\label{table:instLevel}}
\setlength{\tabcolsep}{2.3pt}
\begin{tabular}{l|ccc|c|ccccccc}
 & \multicolumn{3}{c|}{inst meta} & & \multicolumn{7}{|c}{predictions}\\
\# & \textit{k}AN & MV & \dots & act & BP\_1/1 & \dots & BP\_N/1 & \dots & BP\_N/M & C4.5\_1/1 & \dots \\
\hline
\\[-6pt]
77 & 0.92 & 0 & \dots & 2 & 3 & \dots & 2 & \dots & 2 & 3 & \dots\\
\vdots & \vdots & \vdots & & \vdots & \vdots & & \vdots & & \vdots & \vdots &\\
\end{tabular}
\end{table}

At the data set level, several meta-data sets are provided:
\begin{itemize}
 \item a general meta-data set that stores the data set meta-features and the average $N$ by 10-fold cross-validation accuracy for all of the data sets from a learning algorithm with a given hyperparameter setting.
 \item for each learning algorithm a meta-data set stores the data set meta-features, the learning algorithm hyperparameter settings, and the average $N$ by 10-fold cross-validation accuracy for all of the data sets for the given hyperparameter setting.
\end{itemize}

The structure for the general meta-data set is provided in Table \ref{table:dsLevel}.
The structure and information of this meta-data set is typical of that used in previous meta-learning studies that provides a mapping from data set meta-features to accuracies obtained by a set of learning algorithms.
Most previous studies have been limited to only using the default hyperparameters.
The MLRR includes the accuracies from multiple hyperparameter settings.
The hyperparameter settings from each learning algorithm are denoted by a ``LA\_\#'' where LA refers a learning algorithm and \# refers which hyperparameter setting was used for that learning algorithm.

\begin{table}
\centering
{\caption{The structure of the meta-data set at the data set level.}\label{table:dsLevel}}
\setlength{\tabcolsep}{2pt}
\begin{tabular}{l|ccc|cccccc}
 & \multicolumn{3}{c}{data set meta-features} & \multicolumn{6}{|c}{LA accuracies}\\
data set & numInst & numAttr & \dots & BP\_1 & BP\_2 & \dots & BP\_N & C4.5\_1 & \dots\\ 
\hline
\\[-6pt]
iris & 150 & 4 & \dots & 96.80 & 95.07 & \dots & 93.47 & 95.60 & \dots \\
abalone & 4177 & 8 & \dots & 20.27 & 29.84 & \dots & 21.91 & 23.24 & \dots \\ 
\vdots & \vdots & \vdots &  & \vdots & \vdots & \vdots & \vdots & \vdots\\
\end{tabular}
\end{table}

The meta-data sets for each learning algorithm are designed to aid in algorithmic hyperparameter estimation, i.e. given a data set, can we predict which hyperparamter settings will give the highest classification accuracy.
For each learning algorithm, a meta-data set is provided that contains the data set meta-features, the toolkit that was used, the hyperparameter setting and the average accuracy for each unique tool kit/hyperparameter combination.
The structure of the meta-data set for each learning algorithm is provided in Table \ref{table:LA}.
The accuracy (``acc'') represents the average accuracy for all $k$-fold validation runs (i.e. multiple runs of the same learning algorithm with different random seeds to partition the folds).
The toolkit is also provided to allow a user to compare toolkits or only do hyperparameter estimation for a single toolkit.

\begin{table}
\centering
{\caption{The structure of the table for mapping learning algorithm hyperparameters between toolkits.}\label{table:LA}}
\setlength{\tabcolsep}{4.6pt}
\begin{tabular}{l|ccc|c|ccc|c}
 & \multicolumn{3}{c|}{DS meta features} & toolkit & \multicolumn{3}{c|}{hyperparamters} & \\
data set & numInst & numAttr & \dots & weka &  LR & Mo & \dots &  acc\\ 
\hline
\\[-6pt]
iris & 150 & 4 & \dots & weka & 0.71 & 0.61 & \dots & 96.80 \\
iris & 150 & 4 & \dots & weka & 0.11 & 0.25 & \dots & 97.04 \\
\vdots & \vdots & \vdots & & \vdots & \vdots & \vdots & & \vdots\\
\end{tabular}
\end{table}

MLRR provides easy access to researchers and practitioners to a large and varying set of meta-data information as shown in the tables above.
The provided meta-data sets are a snapshot of an underlying database that stores all of the previous experimental results which can be updated as more results are obtained.
A revision history of the data sets is provided so that results can be compared even if the meta-data set has been updated.

\section{DATABASE DESCRIPTION}
\label{section:DBdescription}
MLRR uses MongoDB as the database to store the results from machine learning experiments.
MongoDB is a NoSQL database that allows for adding new features (such as new learning algorithms and/hyperparameters), thus, escaping the rigidity of the more traditional SQL databases.
This allows for expanding the database with new learning algorithms and/or hyperparameters.
In traditional relational databases the information that need to be stored has to known in advance.
When new features are desired, a new schema needs to be created and then the database has to be migrated over to the new schema.
Thus, with a NoSQL database, new learning algorithms/hyperparameters and other pieces of information can easily be added into the MLRR.

The data is stored as a document database as collections of key-value pairs. 
Each collection represents the experimental results on a particular data set.
In each collection, the keys are LA\_hyperparameterSetting.
The value then is a JSON text document that stores the results of an experiment (e.g. the results of 10-fold cross-validation on the iris data set using C4.5).
These documents also contain pointers to other documents that hold information about training/testing sets for each experiment.
The data set/instance level metafeatures are stored in separate documents in their respective data set collection. 
A separate collection stores information about the learning algorithms and their hyperparameters.

The best way to visualize the database is as a heirarchy of key-value pairs as shown in Figure \ref{figure:database}.
At the top-level, there are collections - these are the individual data sets in the database.
Each of them hold a collection of documents that represent an output file, or experiment, named by its learning algorithm with two numbers that correspond to the random seed used to partition the data and the hyperparameter setting.
In these documents, the predictions for each instance is stored.
Collections for which instances were used for training hyperparameter settings are also included.

%

\section{EXTENDING THE DATABASE}
\label{section:extendingDB}
The data provided by MLRR only contains a snapshot of current machine learning results.
To allow more machine learning results to be added and to allow the MLRR to evolve as the state of machine learning evolves, MLRR provides a method to upload new machine learning results.
Currently, scripts are provided to upload the output from running WEKA.
This provides a simple way to upload experimental results from a commonly used toolkit.
The file is slightly modified such that the first line provides which learning algorithm and hyperparameters were used.
The database will have the ability to upload files generated by other toolkits in the future.

\section{INCLUDED META-FEATURES}
\label{section:details}
In this section, we detail the meta-features that are included in the machine learning results repository (MLRR).
We store a set of data set meta-features that have been commonly used in previous meta-learning studies.
Specifically, we used the meta-features from Brazdil et al. \cite{Brazdil2003}, Ho and Basu \cite{Ho2002}, Pfahringer et al. \cite{Pfahringer2000}, and Smith et al. \cite{Smith2012_IH}.
As the underlying database is a NoSQL database, additional meta-features can be easily added in the future.
We now describe the meta-features from each study.

The study by Brazdil et al. \cite{Brazdil2003} examined ranking learning using instance based learning.
The meta-features are designed to be quickly calculated and to represent properties that affect algorithm performance.
\begin{itemize}
 \item \textit{Number of examples}.
This feature helps identify how scalable an algorithm is based the size of its input.
 \item \textit{Proportion of symbolic attributes}.
This feature can be used to consider how well an algorithm deals with symbolic or numeric attributes.
 \item \textit{Proportion of missing values}.
This features can be used to consider how robust an algorithm is to incomplete data.
 \item \textit{Proportion of attributes with outliers}.
An attribute is considered to have an outlier if the ratio of variances of the mean value and the $\alpha$-trimmed mean is smaller than 0.7 where $\alpha = 0.05$.
This feature can be used to consider how robust an algorithm is to outlying numeric values.
 \item \textit{Entropy of classes}.
This feature measures one aspect of problem difficulty in the form of if one class outnumbers another.
\end{itemize}

Ho and Basu \cite{Ho2002} sought to measure the complexity of a data set to identify areas of the data set that contribute to its complexity focusing on the geometrical complexity of the class boundary.
\begin{itemize}
 \item Measures of overlap of individual feature values:
 \begin{itemize}
  \item \textit{The maximum Fisher's Discriminant ratio}.
This is the Fisher's discriminant ratio for an attribute:
$$
f = \frac{(\mu_1 - \mu_2)^2}{\sigma_1^2 + \sigma_2^2},
$$
where $\mu_i$ and $\sigma_i^2$ represent that mean and variance for a class.
The maximum Fisher's discriminant value over the attributes is used for this measure.
For multiple classes, this measure is expanded to:
$$
f = \frac{\sum_{i=1}^C\sum_{j=i+1}^Cp_ip_j(\mu_i - \mu_j)^2}{\sum_{i=1}^Cp_i\sigma_i^2}
$$
where $C$ is the number of classes and $p_i$ is the proportion of instances that belong to the $i^{th}$ class.
  \item \textit{The overlap of the per-class bounding boxes}.
This feature measures the overlap of the tails of the two class-conditional distributions.
For data sets with more than 2-classes, the overlap of the per-class bounding boxes is computed for each pair of classes and the sum of all pairs of classes is returned.
  \item \textit{The maximum (individual) feature efficiency}.
This feature measures how discriminative a single feature is.
For each attribute, the ratio of instances with differing classes that are not in the overlapping region is returned
The attribute that produces the largest ratio of instances is returned.
  \item \textit{The collective feature efficiency}.
This measures builds off of the previous one.
The max ratio is first calculated as before.
Then, the instances that can be discriminated are removed and the maximum (individual) feature efficiency is recalculated with the remaining instances.
This process is repeated until no more isntances can be removed.
The ratio of instances that can be discriminated is returned.
 \end{itemize}
 \item Measures of class separability:
 \begin{itemize}
  \item \textit{The minimized sum of the error distance of a linear classifier}.
This feature measures to what extent training data is linearly separable and returns the difference between a linear classifier and the actual class value.
  \item \textit{The training error of a linear classifier}.
This feature also measures to what extent the training training data is linearly separable.
  \item \textit{The fraction of points on the class boundary}.
This feature estimates the length of the class boundary by constructing a minimum spanning tree over the entire data set and returning the ratio of the number of nodes in the spanning tree that are connected and belong to different classes to the number of instances in the data set.
  \item \textit{The ratio of average intra/inter class nearest neighbor distance}.
This measure compares the with-in class spread with the distances to the nearest neighbors of the other classes.
For each instance, the distance to its nearest neighbor with the same class ($intraDist(x)$) and to its nearest neighbor with a different class ($interDist(x)$) is calculated.
Then the measure returns:
$$
\frac{\sum_i^N intraDist(x_i)}{\sum_i^N interDist(x_i)}
$$
where $N$ is the number of instances in the data set.
  \item \textit{The leave-one-out error rate of the one-nearest neighbor classifier}.
This feature measures how close the examples of different classes are.
 \end{itemize}
 \item Measures of geometry, topology, and density of manifolds
 \begin{itemize}
  \item \textit{The nonlinearity of a linear classifier}.
Following Hoekstra and Duin \cite{Hoekstra1996}, given a training set, a test set is created by linear interpolation with random coefficients between pairs of randomly selected instances of the same class.
The error rate of a linear classifier trained with the original training set on the generated test set is returned.
  \item \textit{The nonlinearity of the one-nearest neighbor classifier}.
A test set is created as with the previous feature, but the error rate of a 1-nearest neighbor classifier is returned.
  \item \textit{The fraction of maximum covering spheres}.
A covering sphere is created by centering on an instance and growing as much as possible before touching an instance from another class.
Only the largest spheres are considered.
The measure returns the number of spheres divided by the number of instances in the data set and provides an indication of how much the instances are clustered in hyperspheres or distributed in thinner structures.
  \item \textit{The average number of points per dimension}.
This measure is the ratio of instances to attributes and roughly indicates how sparse a data set is.
 \end{itemize}
\end{itemize}
Multi-class modifications are made according to the implementation of the data complexity library (DCoL) \cite{DCoL}.

Pfahringer et al. \cite{Pfahringer2000} introduced the notion of using performance values (i.e. accuracy) of simple and fast classification algorithms as meta-features.
The landmarkers that are included in the MLRR are listed below.
\begin{itemize}
 \item \textit{Linear discriminant learner}.
Creates a linear classifier that finds a linear combination of the features to separate the classes.
 \item \textit{One nearest neighbor learner}.
Redundant with the leave-one-out error rate of the one-nearest neighbor classifier from Ho and Basu.
 \item \textit{Decision node learning}.
A decision stump that splits on the attribute that has the highest information gain.
A decision stump is a decision tree with only one node.
 \item \textit{Randomly chosen node learner}.
A decision stump that splits on a randomly chosen attribute.
 \item \textit{Worst node learner}.
A decision stump that splits on the attribute that has the lowest information gain.
 \item \textit{Average node learner}.
A decision stump is created for each attribute and the average accuracy is returned.
\end{itemize}
The use of landmarkers have been shown to be competitive with the best performing meta-features with a significant decrease in computational effort \cite{Reif2014}.

Smith et al. \cite{Smith2012_IH} sought to identify and characterize instances that are difficult to classify correctly.
The difficulty of an instance was determined based on how frequently it was misclassified.
To characterize why some instances are more difficult than others to classify correctly, the authors used different hardness measures.
They include:
\begin{itemize}
 \item \textit{$k$-Disagreeing Neighbors}.
The percentage of $k$ nearest neighbors that do not share the target class of an instance.
This measures the local overlap of an instance in the original space of the task.
 \item \textit{Disjunct size}.
This feature indicates how tightly a learning algorithm has to divide the task space to correctly classify an instance.
It is measured as the size of a disjunct that covers an instance divided by the largest disjunct produced, where the disjuncts are formed using the C4.5 learning algorithm.
 \item \textit{Disjunct class percentage}.
This features measure the overlap of an instance on a subset of the features.
Using a pruned C4.5 tree, the disjunct class percentage is the number of instances in a disjunct that belong to the same class divided by the total number of instances in the disjunct.
 \item \textit{Tree depth (pruned and unpruned)}.
Tree depth provides a way to estimate the description length, or Kolmogorov complexity, of an instance.
It is the depth of the leaf node that classifies an instance in an induced tree.
 \item \textit{Class likelihood}.
This features provides a global measure of overlap and the likelihood of an instance belonging to the target class.
It is calculated as:
$$
\prod_i^{|x|} p(x_i|t(x))
$$
where $|x|$ represents the number of attributes for the instance $x$ and $t(x)$ is the target class of $x$.
 \item \textit{Minority value}.
This feature measures the skewness of the class that an instance belongs to.
It is measured as the ratio of instances sharing the target class of an instance to the number of instances in the majority class.
 \item \textit{Class balance}.
This feature also measures the class skew.
First, the ratio of the number of instances belonging the target class to the total number of instances is calculated.
The difference of this ratio with the ratio of one over the number of possible classes is returned. If the class were completely balanced (i.e. all class had the same number of instances), a value of 0 would be returned for each instance.
\end{itemize}
The hardness measures are designed to capture the characteristics of why instances are hard to classify correctly.
Data set measures can be generated by averaging the hardness measures over the instances in a data set.

\section{CONCLUSIONS AND FUTURE WORK}
\label{section:conlcusions}
In this paper, we presented the \textit{machine learning results repository} (MLRR) an easily accessible and extensible database for meta-learning.
MLRR was designed with the main goals of providing an easily accessible data repository to facilitate meta-learning and providing benchmark meta-data sets to compare meta-learning experiments.
To this end, the MLRR provides ready to download meta-data sets of previous experimental results.
MLRR is unique in that it also provides meta-data at the instance level.
Of course, the results could also be used as a means of comparing one's work with prior work as they are stored in the MLRR.
The MLRR can be accessed at \url{http://axon.cs.byu.edu/mlrr}.

The MLRR allows for reproducible results as the data sets are stored on the server and as the class names and toolkits are provided.
The experiment DB is a lot more rigid in its design as it is based on relational databases and PMML (predictive model markup language), thus, requiring a learning curve to import and extract data.
The MLRR is less rigid in its design allowing for easier access to the data and more extensibility, with the trade-off of less formality.

One direction for future work is to integrate the API provided at openml\footnote{\url{www.openml.org}} (an implementation of an experiment database) to incorporate their results with those that are in the MLRR.
This will help provide easy access to the results that are already stored in openml without having to incur the learning cost associated with understanding the database schema.
Another open problem is how to store information about how a data set is preprocessed.
Currently, the MLRR can store the instance level information resulting from preprocessing, but it lacks a mechanism to store the preprocessing process.
Integrating this information in an efficient way is a direction of current research.


\begin{landscape}
\begin{figure}[hpt]
\centering
\begin{tikzpicture}[
    grow=down,
    level 1/.style={sibling distance=3.6cm,level distance=3cm},
    level 2/.style={sibling distance=0.6cm, level distance=6cm},
    edge from parent/.style={very thick,draw=blue!40!black!60,
        shorten >=5pt, shorten <=5pt},
    edge from parent path={(\tikzparentnode.south) -- (\tikzchildnode.north)},
    kant/.style={text width=2cm, text centered, sloped},
    every node/.style={text centered, inner sep=2mm},
    punkt/.style={rectangle, rounded corners, shade, top color=white,
    bottom color=blue!50!black!20, draw=blue!40!black!60, very
    thick }
    ]

\node[punkt, text width=2.5em, align=center] {Root}
    child {
        node[punkt, text width=2em] {iris}
        child[left=1cm] {
            node [punkt,rectangle split, rectangle split,
            rectangle split parts=4] {
                \textbf{fold:\{\#:Pred\}}
                \nodepart{second}
                1\{1:1, 57:2, \dots\}
                \nodepart{third}
                2\{2:1, 107:3, \dots\}
                \nodepart{fourth}
                \dots
            }
            edge from parent
                node[above, kant,  pos=0.85] {BP\_1}
        }
        child[left=0.75cm] {
            node [punkt, rectangle split, rectangle split parts=4]{
                \textbf{fold:\{\#:Pred\}}
                \nodepart{second}
                1\{15:1, 147:3, \dots\}
                \nodepart{third}
                2\{26:1, 67:2, \dots\}
                \nodepart{fourth}
                \dots
            }
            edge from parent
                node[kant, above, pos=0.85] {BP\_2}
	      }
        child[left=0.4cm] {
            node{
                \begin{LARGE}\dots\end{LARGE}
            }
	      }
        child[sibling distance=0.4cm] {
            node [punkt, rectangle split, rectangle split parts=4]{
                \textbf{MF:val}
                \nodepart{second}
                kAN:0.97
                \nodepart{third}
                DS:0.84
                \nodepart{fourth}
                \dots
            }
            edge from parent
                node[kant, above, pos=0.8] {meta features}
	      }
}
    child {
        node[punkt, text width=3em] {abalone}
        child[right=-0.9cm] {
            node [punkt,rectangle split, rectangle split,
            rectangle split parts=4] {
                \textbf{fold:\{\#:Pred\}}
                \nodepart{second}
                1\{1035:9, 7:2, \dots\}
                \nodepart{third}
                2\{9:1, 237:3, \dots\}
                \nodepart{fourth}
                \dots
            }
            edge from parent
                node[above, kant,  pos=.85] {BP\_1}
        }
        child[left=-2.15cm] {
            node [punkt, rectangle split, rectangle split parts=4]{
                \textbf{fold:\{\#:Pred\}}
                \nodepart{second}
                1\{15:1, 147:3, \dots\}
                \nodepart{third}
                2\{26:1, 67:2, \dots\}
                \nodepart{fourth}
                \dots
            }
            edge from parent
                node[kant, above, pos=0.85] {BP\_2}
	      }
        child[sibling distance=4.7cm] {
            node{
                \begin{LARGE}\dots\end{LARGE}
            }
	      }
        child[sibling distance=2.4cm] {
            node [punkt, rectangle split, rectangle split parts=4]{
                \textbf{MF:val}
                \nodepart{second}
                kAN:0.24
                \nodepart{third}
                DS:0.14
                \nodepart{fourth}
                \dots
            }
            edge from parent
                node[kant, above, pos=0.8] {meta features}
	      }
}
     child {
        node{\begin{Huge}\dots\end{Huge}}
    }
    child {
        node[punkt, text width=3em] {training sets}
        child[left=-1.2cm] {
            node [punkt,rectangle split, rectangle split,
            rectangle split parts=4] {
                \textbf{seed\_numFolds\_fold}
                \nodepart{second}
                1\_10\_1\{1:1, 2:?, \dots\}
                \nodepart{third}
                1\_10\_2\{1:1, 2:1, \dots\}
                \nodepart{fourth}
                \dots
            }
            edge from parent
                node[above, kant,  pos=.85] {weka}
        }
        child[left=-1.3cm] {
            node [punkt,rectangle split, rectangle split,
            rectangle split parts=4] {
                \textbf{seed\_numFolds\_fold}
                \nodepart{second}
                1\_10\_1\{1:1, 2:?, \dots\}
                \nodepart{third}
                1\_10\_2\{1:1, 2:1, \dots\}
                \nodepart{fourth}
                \dots
            }
            edge from parent
                node[above, kant,  pos=.85] {waffles}
	      }
         child[right=0.8cm]{
            node{
                   \begin{LARGE}\dots\end{LARGE}
                }
              }
    }
    child {
        node[punkt, text width=6.5em, right=0.4cm] {hyperparameters}
        child[sibling distance=1cm] {
            node [punkt, rectangle split, rectangle split parts=4]{
                \textbf{\#:HP setting}
                \nodepart{second}
                1\{Co:0.25,Pr:1,\dots\}
                \nodepart{third}
                2\{Co:0.1,Pr:0,\dots\}
                \nodepart{fourth}
                \dots
            }
            edge from parent
                node[kant, above, pos=0.85] {C4.5}
	      }
        child {
            node [punkt,rectangle split, rectangle split,
            rectangle split parts=4] {
                \textbf{\#:HP setting}
                \nodepart{second}
                1\{LR:0.26,Mo:0.16\}
                \nodepart{third}
                2\{LR:0.26,Mo:0.15\}
                \nodepart{fourth}
                \dots
            }
            edge from parent
                node[above, kant,  pos=.9] {BP}
        }
        child[sibling distance=2.1cm] {
            node{
                \begin{LARGE}\dots\end{LARGE}
            }
	      }
}
;
\end{tikzpicture}
\caption{Hierarchical representation of how the results from machine learning experiments are stored in the NoSQL database for the MLRR.
Each data set has a collection containing the predictions for each instance from a learning algorithm as well as its meta-features.
A separate collection stores all of the information for the learning algorithms and which hyperparameters were used.
Another collection stores the information for which instances were used for training.} \label{figure:database}
\end{figure}
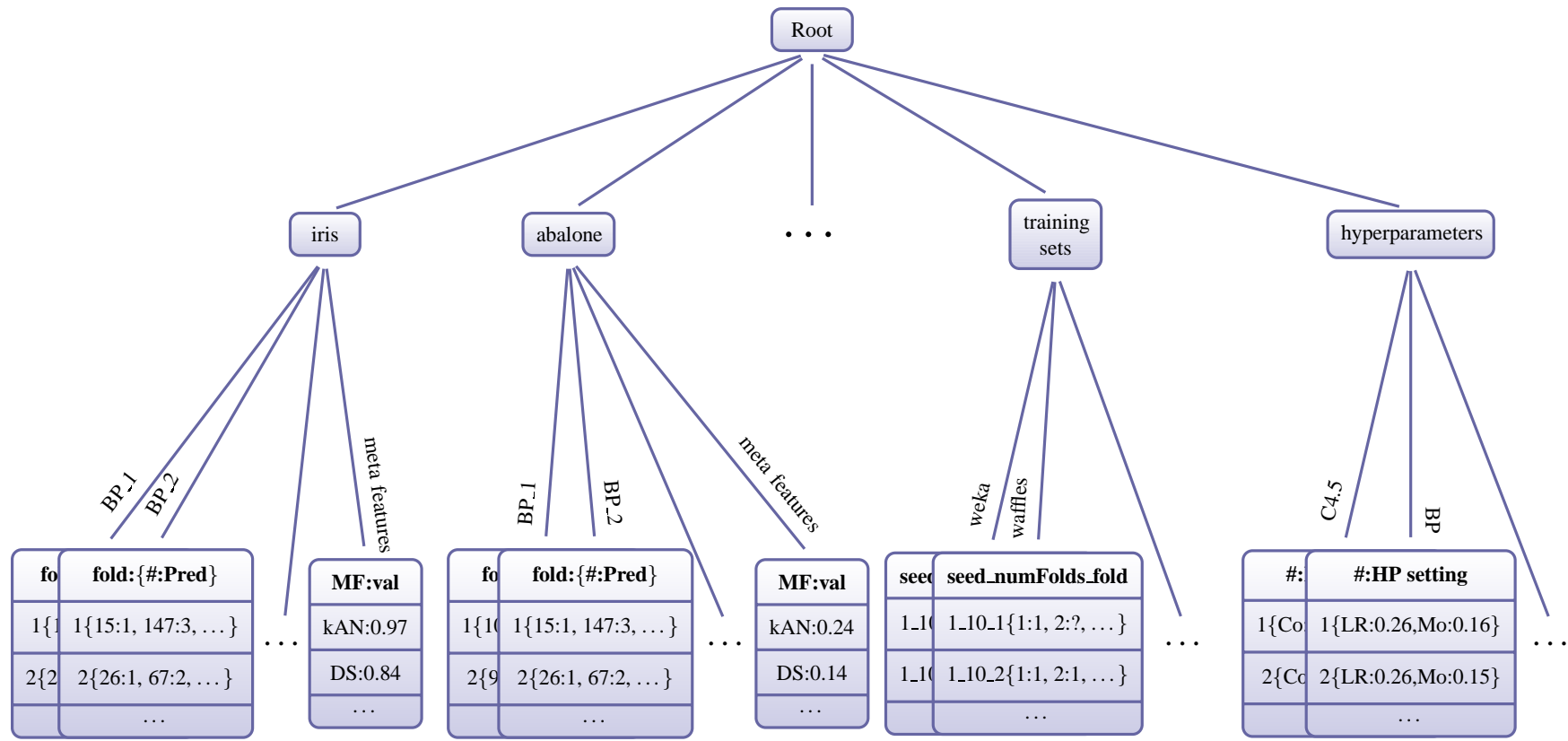
\end{landscape}


\end{document}